\documentclass[lettersize,journal]{IEEEtran}
\usepackage{amsmath,amsfonts,amssymb,amsthm}
\usepackage{algorithmicx}
\usepackage{algpseudocode}
\usepackage{algorithm}
\usepackage{array}
\usepackage[caption=false,font=normalsize,labelfont=sf,textfont=sf]{subfig}
\usepackage{textcomp}
\usepackage{stfloats}
\usepackage{url}
\usepackage{verbatim}
\usepackage{graphicx}
\usepackage{todonotes}
\usepackage{listings}
\usepackage{calrsfs}
\usepackage{tikz,pgfplots}
\usepackage[space]{cite}

\usepackage[style=italian]{csquotes}
\SetBlockThreshold{0}

\newcommand\gptblockquote[3]{
\begin{quotation}
\small
{\textbf{#1}#2}\hfill(#3)
\end{quotation}
\par}

\begin{document}
\title{Microscaling Floating Point Formats\\
for Large Language Models}
 \author{
      Marco~Cococcioni$^{1*}$,
      Dario~Pagani$^1$, and
      Federico~Rossi$^1$
 \thanks{$^1$Marco Cococcioni, Dario Pagani, Federico Rossi, are with the University of Pisa, Department of Information Engineering, Pisa, Italy (e-mail: marco.cococcioni@unipi.it, dario.pagani.146@gmail.com, federico.rossi@ing.unipi.it)
 \\$^*$corresponding author}
}

\markboth{IEEE Transactions on Computers, under review, 2025}%
{Cococcioni et al. \MakeLowercase{\textit{et al.}}: Microscaling Floating Point Formats for Large Language Models}


\maketitle

\begin{abstract}
The increasing computational and memory demands of large language models (LLMs) necessitate innovative approaches to optimize resource usage without compromising performance. This paper leverages microscaling floating-point formats, a novel technique designed to address these challenges by reducing the storage and computational overhead associated with numerical representations in LLMs. Unlike traditional floating-point representations that allocate a dedicated scale for each value, microscaling employs a shared scale across a block of values, enabling compact one-byte floating-point representations while maintaining an extended dynamic range. We explore the application of microscaling in the context of 8-bit floating-point formats to significantly reduce memory footprint and computational costs. We tested several configurations of microscaling floats within the GPT-2 LLM architecture, demonstrating that microscaling data formats can achieve competitive accuracy during training and inference, proving its efficacy as a resource-efficient alternative for deploying LLMs at scale.
The source code is publicly available at: \url{https://github.com/unipi-dii-compressedarith/llm.c-sve} . C++23 or a higher standard is required to compile it. 
\end{abstract}

\begin{IEEEkeywords}
Microscaling, floating-point, C++, LLM, minifloat, IEEE 754 standard.
\end{IEEEkeywords}

\section{Introduction}

The rapid advancement in the size and complexity of large language models (LLMs) has revolutionized applications in natural language processing and beyond. However, the increasing computational and memory demands associated with these models present substantial challenges, particularly in terms of scalability and cost-efficiency. For instance, models such as GPT-2 (Generative Pre-Trained Transformer, version 2) and its successors have highlighted the need for innovative strategies to reduce resource consumption without sacrificing performance \cite{rouhani2023microscalingdataformatsdeep, radford2019language}.

A crucial aspect of this challenge lies in the numerical representation of model parameters, which directly impacts storage and computational requirements. Traditional floating-point formats, including half-precision and BFLOAT16, have been widely adopted in deep learning due to their balance between precision and efficiency \cite{kalamkar2019studybfloat16deeplearning}. However, as models grow larger, these formats become increasingly insufficient, prompting the exploration of alternative approaches.

One promising solution is Microscaling (MX) floating-point formats, a technique that leverages shared scaling across blocks of values to enable compact numerical representations. Unlike conventional formats, which allocate individual exponents to each value, Microscaling reduces memory usage by representing values with just one byte and applying a shared exponent. This approach allows for significant savings in memory and computation while maintaining a dynamic range suitable for deep learning tasks \cite{rouhani2023microscalingdataformatsdeep}.

This study focuses on 8-bit floating-point formats within the Microscaling framework, a choice motivated by recent advancements in hardware support for low-bit representations. By reducing the memory footprint and computational overhead of LLMs, Microscaling has the potential to enable more efficient training and deployment, particularly in resource-constrained environments. This research aims to evaluate the trade-offs between accuracy and efficiency inherent in Microscaling, contributing to the broader goal of sustainable and scalable AI development.

\subsection{Main novelties of the work and obtained insights}

In this work we provide a fast and modern C++ implementation of a generative pretrained transformer, equipped, for the first time, with the recently introduced microscaling format.


In addition, we introduce important mitigations needed to accurately train the GPT model using a mixed-precision approach, where different numeric formats are used in the appropriate context, to deliver an acceptable level of performance while reaching a lower bandwidth consumption and memory footprint.

The main obtained insights, that can be useful to other researcher and practitioners working in this domain, are:
\begin{enumerate}
\item The operation of tensor transposition is not exactly trivial in Microscaling; while this may seem self-evident in retrospect, it is not entirely trivial to discern at the beginning
\item When rounding numbers whose significands are stored on a tiny number of bits it is vital to use an appropriate rounding policy instead of relying on the faster truncation
\item The importance of correctly ordering the operations on low precision formats to reduce numerical error
\end{enumerate}

\subsection{Paper organization}
This paper is organized as follows: Section~\ref{sec:related} provides an overview of related works, discussing existing numerical representations and their impact on deep learning efficiency; this is followed by a detailed explanation of Microscaling formats in Section \ref{sec:mx}, including their theoretical foundation and implementation details; Section~\ref{sec:impl} and~\ref{sec:fp-mod} present the implementation details of Microscaling within the LLM.c GPT-2 library, outlining the datasets, models, and evaluation metrics used; Section~\ref{sec:eval} ("Evaluation and results") analyzes the performance of Microscaling, focusing on memory efficiency, computational throughput, and accuracy; finally, Section~\ref{sec:concl}, devoted to conclusions, highlights the implications of these findings, and 
suggests directions for future research.

\section{Related works} \label{sec:related}

The challenge of optimizing numerical representations in machine learning has received significant attention in recent years. Traditional approaches such as single and half-precision floating-point formats have been instrumental in enabling the rapid growth of deep learning. BFLOAT16, in particular, has been widely adopted for its compatibility with existing hardware and its ability to balance precision with computational efficiency \cite{kalamkar2019studybfloat16deeplearning}.

Emerging research has focused on reducing the bit-width of numerical representations further, with 8-bit floating-point formats gaining traction due to their promise of improved memory efficiency. Hardware vendors have begun incorporating support for these formats, paving the way for their adoption in large-scale AI systems \cite{rouhani2023microscalingdataformatsdeep}.

Another line of work explores alternative architectures and methods to enhance computational efficiency in LLMs. The introduction of attention mechanisms, as demonstrated in Transformer-based models, revolutionized deep learning by offering a scalable approach to sequence modeling \cite{vaswani2023attentionneed}. However, these advancements also exacerbated the resource demands of training and inference, highlighting the need for complementary techniques like Microscaling to address the bottlenecks associated with memory and computation.

The concept of shared scaling, as implemented in Microscaling formats, represents a novel approach to numerical representation. It extends the principles of block quantization by introducing a shared exponent, enabling finer-grained memory optimization without compromising the model's ability to capture complex patterns \cite{rouhani2023microscalingdataformatsdeep}. This study builds on these foundations, aiming to further validate the effectiveness of Microscaling in large language models.

\section{Microscaling Floating-Point Formats} \label{sec:mx}

Floating point arithmetic uses integral numbers to represent
a subset of rational numbers and is a way used to represent real
numbers in the computer \cite{10.1145/103162.103163}.
With such an approach, a number is represented by three fields:
the sign bit, the exponent and the significand (also known as the mantissa).

\begin{figure}[h]
    \centering
    \includegraphics[width=\linewidth]{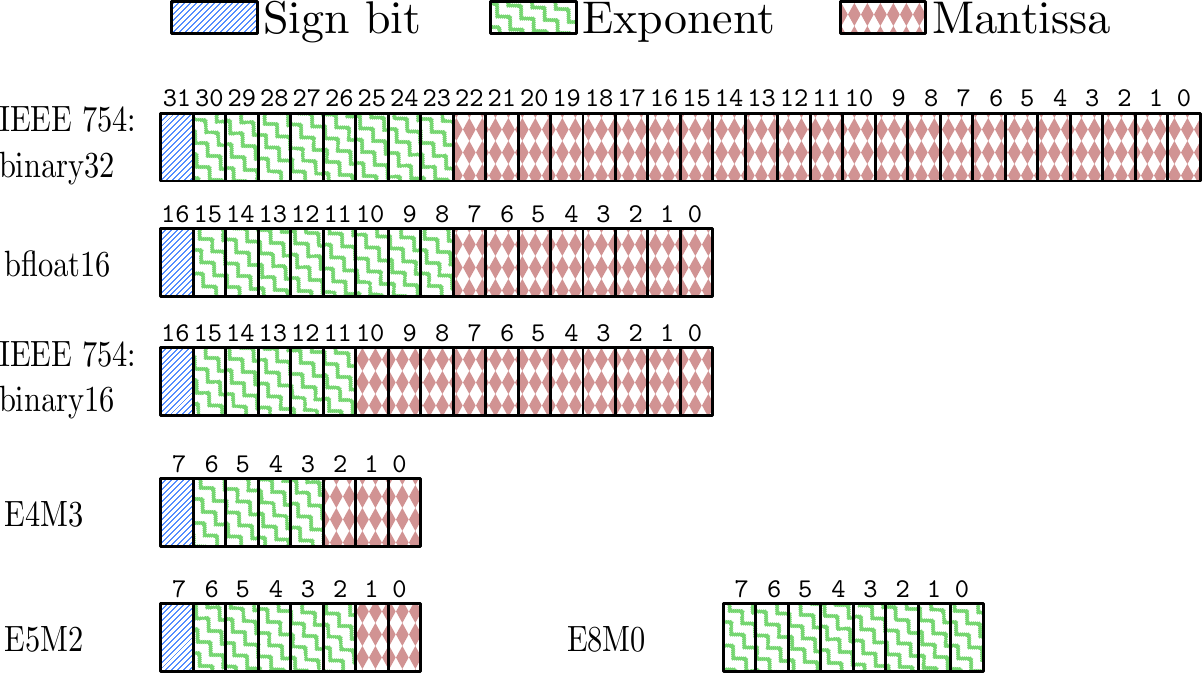}
    \caption{Comparison between some floating point formats}
    \label{fig:float-cmp}
\end{figure}

In addition to rational numbers, these formats can represent the special values of

\begin{itemize}
    \item \texttt{NaN}, typically used to signal an error in a procedure, for instance the computation of the square root of a negative number $\sqrt{-1}$;
    \item two signed infinities $\pm \infty$, typically used to signal an overflow in the computation, for instance $\ln\left(+0\right)$;
    \item and two signed zeros $\pm 0$.
\end{itemize}

The Microscaling format aims to overcome the loss of
dynamic range when using 8 bits data formats to represent
real numbers --- such as E4M3 and E5M2 --- this is 
done by organizing a given set  of numbers into small blocks 
and associating to each block a scale --- also called weight, --- 
each block element is then scaled, converted and
stored using a small data format,
typically on 8 bits \cite{rouhani2023microscalingdataformatsdeep, ocp_mx}. 

\subsection{Sofware-defined floating points}

Since it is uncommon for present-day hardware to have support for eight bit floating-point numerical formats, we had to make use of software emulation. In particular we used a modified version of the
\texttt{anyfloat} class from the 
cppposit library \cite{posit}

Anyfloat is a template class that allows the programmer to define custom floating-point numeric types
with an arbitrary number of bits for their significands and 
exponents, most operations are performed by converting
the number back and forth between the custom representation
and the native type handled by the machine's FPU --- such as
single precision, double precision or even half-precision floats on
certain newer machines.

\begin{equation}
\label{formula:float_form}
x = \begin{cases}
    -1^s \cdot \infty &   \text{if} \; k = 2^E - 1 \land m = 0\\

    \texttt{NaN} &  \text{if} \; k = 2^E - 1 \land m \neq 0 \\

    \pm 0 &   \text{if} \; k = 0 \land m = 0\\

    -1^s \cdot 2^{-b + 1} \cdot \dfrac{m}{2^M - 1} & \text{if} \; k = 0 \land \texttt{denorm enabled} \\
    
    -1^s \cdot 2^{k - b} \cdot \left( 1 + \dfrac{m}{2^M - 1} \right) & \text{otherwise} \\

    \end{cases}
\end{equation}

Equation~\ref{formula:float_form} describes how to compute
the represented number starting from the values of the three
fields, sign, exponent and mantissa; read as unsigned integral
numbers, noted as $s$, $k$ and $m$ respectively and stored on
$1$, $E$ and $M$ bits respectively. The constant $b$, called the 
exponent's bias, is computed as $b = 2^{E-1} - 1$.
This format is similar and 
binary compatible to the one defined by the IEEE 754 standard.
Support for denormalized numbers (also known as subnormal 
numbers) can be enabled or disabled by the programmer when
defining the custom data type.


\begin{figure}[h]
    \centering
    \includegraphics[width=.5\linewidth]{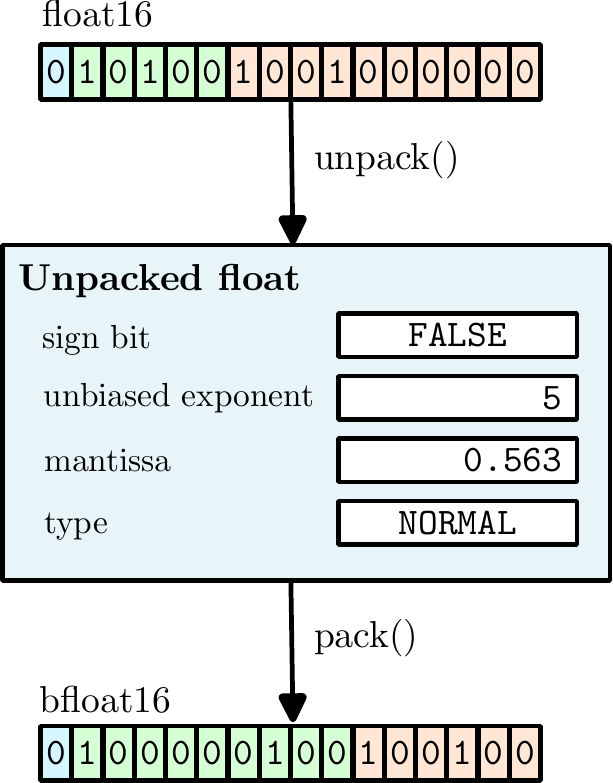}
    \caption{Anyfloat  makes use of a support data structure, called Unpacked, to represent the unpacked number}
    \label{fig:anyfloat-unpacked}
\end{figure}

\subsection{The Microscaling data-format}
\label{sec:mx_format}

The Microscaling format aims to overcome the loss of
dynamic range when using 8 bits data formats to represent
real numbers --- such as E4M3 and E5M2 --- this is 
done by diving a list of numbers into small chunks 
and associating to each chunk a \textit{shared scale},
each chunk's \textit{element} is then scaled, converted and
stored using a small data format,
typically on 8 bits \cite{rouhani2023microscalingdataformatsdeep, ocp_mx}.

Interestingly, it is possible to perform certain mathematical operations without the need to decompress the quantized values; most importantly, the dot-product between two Microscaling vectors can be computed by firstly computing the unscaled intra-block dot product, scaling it properly and finally accumulating the results to compute the inter-block dot product \cite{ocp_mx}.

Let us call $x \in \mathbb{M}_{OP}^n$ the vector of the original unquantized values, $c \in \mathbb{M}_{RP}^n$ the vector of the quantized values (where $\mathbb{M}_{OP} and \mathbb{M}_{RP}$ denote the set of machine numbers in the original and reduced precision respectively), $B$ the block length, $l = \left\lceil \frac{n}{B} \right\rceil$ the number of Microscaling blocks, $\forall i \in \left\{1, \dots, l\right\} \; s_i = 2^{w_i} \land w_i \in \mathbb{Z}$ the blocks' shared scales; then we can rewrite the dot-product between two vectors $x^a, x^b$ as shown in Equation~\ref{formula:dot1}:

\begin{align}
    y &= x^a \cdot x^b = \sum_{j = 1}^{n} x^a_i \cdot x^b_i \nonumber \\
    y &\simeq \sum_{j = 1}^{l}\sum_{i = 1}^{B} s_j^a c_i^a \cdot s_j^b c_i^b \nonumber \\
    y &\simeq \sum_{j = 1}^{l} s_j^a s_j^b \sum_{i = 1}^{B} c_i^a \cdot c_i^b \label{formula:dot1}
\end{align}

\subsection{Microscaling implementation}

We implemented the aforementioned Microscaling data structure
by defining a template class \texttt{microscaling::vector}
with parameters to specify which data format has to be
used to store the elements and the scales, how long
a block shall be and if the auto-commit feature 
is enabled.

Our implementation allows the use of a large variety of numerical formats to represent the quantized values, for instance: natives floating-point types, emulated ones, arbitrary-precision and even novel formats such as Posit.

\paragraph{Data organization}
The container stores the data on the heap in two 
\texttt{std::vector}s, one for the scales and one
for the scaled numbers. To retrieve the original
number it is thus sufficient to multiply the scaled
number with its block's scale. Equation~\ref{formula:scale} shows the decompression
operation.

\begin{equation}
\label{formula:scale}
x_i \simeq \begin{cases}
    c_i \cdot s_{\left \lfloor{\frac{i}{B}}\right\rfloor} &\text{if $s_{\left \lfloor{\frac{i}{B}}\right\rfloor}$ is not \texttt{NaN}}\\

    \texttt{NaN} &\text{otherwise}

    \end{cases}
\end{equation}

It is possible to randomly access data in the vector
through the method \texttt{float operator[](int)}.

\begin{figure}
    \centering
    \includegraphics[width=\linewidth]{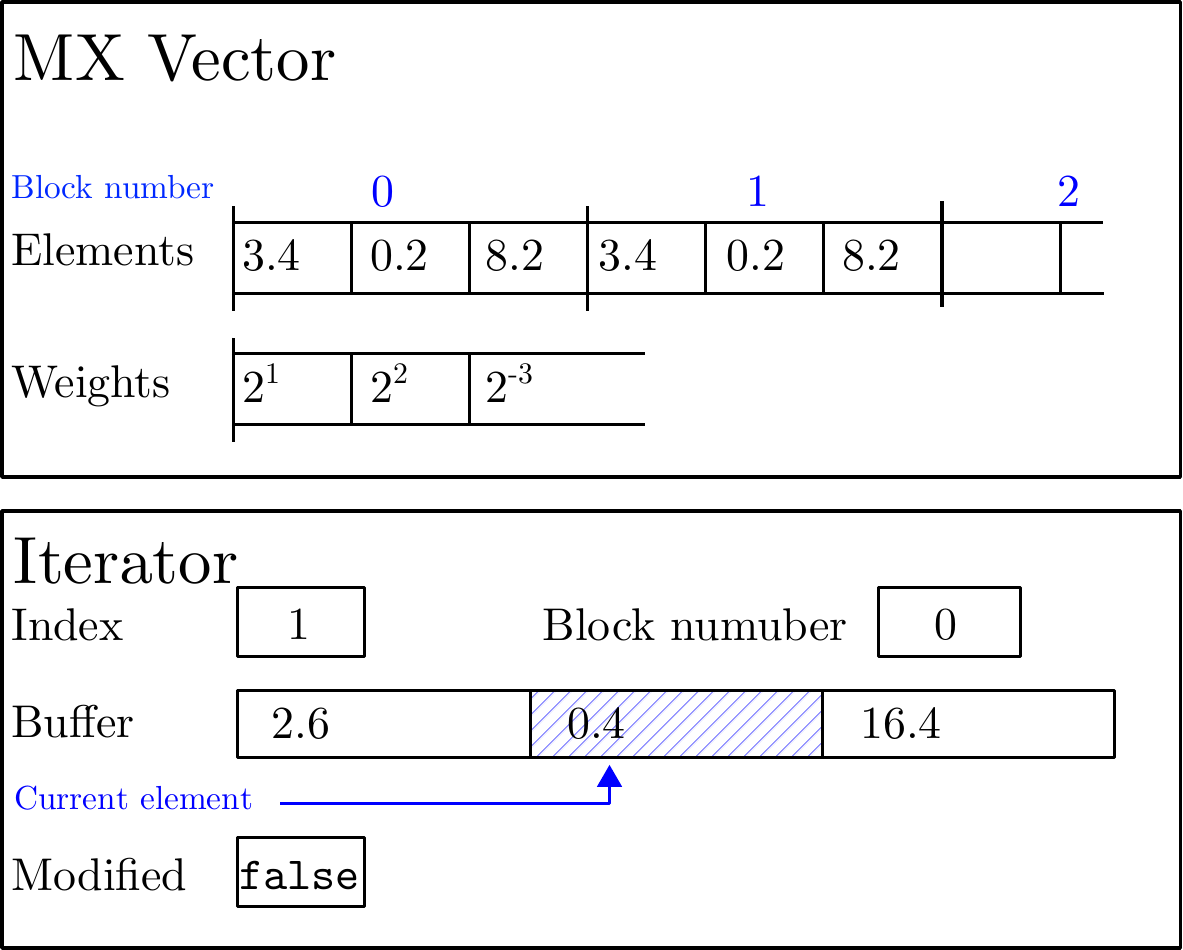}
    \caption{Example of an MX vector with block size 3 and an iterator instance with the auto-commit feature enabled}
    \label{fig:mx_iterator}
\end{figure}

\paragraph{Iterator}
To improve access performance during sequential access as well
as providing writing access to the data structure, we implemented
an \textit{STL}-like iterator.
The Iterator data structure, as shown in Figure~\ref{fig:mx_iterator}, consists of a buffer
that contains the block's uncompressed data.

The buffer is built each time a call to \texttt{operator++()} or
\texttt{operator+(int)} moves the iterator to a new block
or a new iterator is constructed from scratch --- such as 
with a call to \texttt{begin()} on the vector.

The iterator class provides the methods \texttt{refresh()} and
\texttt{commit()} to re-build the buffer and to write
any changes to the buffer back to the main data structure.

\paragraph{Writeable iterators}
We also implemented an optional auto-commit feature,
that is the iterator instance will keep
track of any changes to the buffer during its lifespan and, if
necessary, will automatically make a commit to the main data
structure; notice that the mechanism does not provide any
kind of synchronization between multiple iterators operating
on the same data structure, it is always the programmer's
responsibility to keep the main vector coherent by avoid concurrent accesses. This
feature is mainly to allow the use of \textit{STL}'s algorithms 
such as \texttt{std::copy}, \texttt{std::fill},
\texttt{std::transform} etc...;
as well as allowing for a drop-in replacement for common 
algorithms.

\paragraph{Chunk construction}
When provided with uncompressed data there are several ways
to create a Microscaling vector, we implemented 
the base algorithm mandated by the OCP specification \cite{rouhani2023microscalingdataformatsdeep, ocp_mx} and described in Algorithm~\ref{alg:float2mx}; such a procedure reads a chunk worth of 
data and computes the exponent --- that is the exponent 
field of a floating point number --- of the largest absolute
normal number, then scales all numbers by dividing each of them
by the maximum exponent and finally stores them in the elements'
vector and the scale in the scales' vector.

\begin{algorithm}
\caption{Construction of an MX block}
\label{alg:float2mx}
\begin{algorithmic}[1]
\Require $X = \left\{x_1, \dots, x_B \right\}$
\Require $\xi = $ the largest representable exponent in the quantized data format
\Ensure $s = $ block's shared scale
\Ensure $C = \left\{c_1, \dots, c_B \right\} = $ block's quantized elements 

\If{$\forall x \in X \quad \texttt{not std::isnormal}\left(x_i\right) $} 
\State $w \leftarrow 0$
\Else
\State $p \leftarrow \max\limits_{i = 1\dots B} \left\{ |x_i| : \texttt{std::isnormal}\left(x_i\right) \right\}$\footnotemark
\State $w \leftarrow {
\left\lceil \log_2
    \left( p \right)
\right\rceil } - \xi = \texttt{std::ilogb}\left(p\right) - \xi
$
\footnotemark
\State Clamp $w$ if necessary
\EndIf

\For{$i = 1 \dots B$}
\State $c_i \leftarrow \dfrac{x_i}{2^w} = \texttt{std::scalbn}\left(x_i, -w\right)$
\EndFor

\State $s \leftarrow 2^w$

\end{algorithmic}
\end{algorithm}

Such an approach maximizes the available dynamic range in a block, reducing the risk of certain elements being approximated as zero (underflow).

\section{Implementation} \label{sec:impl}

Now let us introduce the main architectural changes we had to implement to allow the Microscaling data-format to work properly with the GPT-2 network. As a start we used Karpathy's C implementation of GPT-2 \cite{Karpathy2020}, as it was deemed easier to adapt to our C++ library for software emulated floating-point formats. Firstly we ported the project in C++ and divided the source code, previously contained in one source file, into multiple files to make use of translation units for a quicker compilation time, the project was configured using CMake.

\subsection{Generalization}
The base implementation uses C arrays of full-precision floats to store all the values of the network, we modified the program to allow it to use custom containers and different numerical types to represent these values. It is also possible to use heterogeneous types to store the following types of values: weights, activation values, gradients and the buffer for the AdamW optimizer.

\subsection{Weights' master copy}
We implemented this method of learning that consists of keeping a separate copy of the weights in full-precision, called the master copy. During the forward propagation all computations are performed in half-precision --- or with Microscaling, --- but the optimizer, such as Adam or stochastic gradient descent, modifies the master-copy \cite{kalamkar2019studybfloat16deeplearning}.

\begin{figure}[h]
    \centering
    \includegraphics[width=1\linewidth]{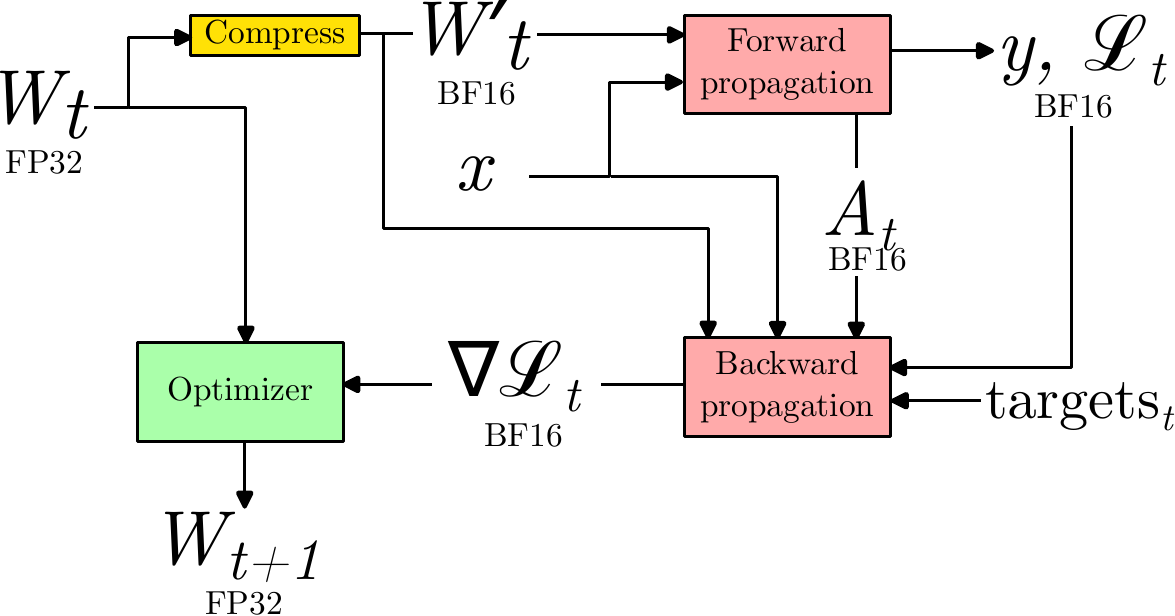}
    \caption{Learning with mixed precision and with a weights' master copy (figure inspired by the one provided in \cite{kalamkar2019studybfloat16deeplearning}).}
    \label{fig:bfloat16_mastercopy}
\end{figure}

\subsection{On-line matrix compression}
As proposed by \cite{rouhani2023microscalingdataformatsdeep}, this method leverages the Microscaling format specifically for the most computationally intensive part of the network, the matrix multiplication. In such an approach the input matrices $W$ and $X$ are compressed with Microscaling and then the dot product is performed on the compressed representations, with promotion to a higher precision format, lastly the bias added. The rest of the network continues to operate using the traditional full or half-precision formats.

\begin{alignat*}{2}
&Y = X \cdot W + b \qquad &&
X \in \mathbb{R}^{B \cdot T \times C} \;
W \in \mathbb{R}^{OC \times C} \\ &&&
b \in \mathbb{R}^{OC} \;
Y \in \mathbb{R}^{B \cdot T \times OC}
\end{alignat*}

During back-propagation, that is when the the program has to compute the gradient of the loss for each layer's input, the gradients can be computed as follows:

\begin{align*}
    \dfrac{\partial{\mathcal{L}}}{\partial{W}} &= X^\intercal \cdot \dfrac{\partial{\mathcal{L}}}{\partial{Y}} \\
    \dfrac{\partial{\mathcal{L}}}{\partial{b}} &=  \left[  \sum_i {\dfrac{\partial{\mathcal{L}_i}}{\partial{y_1}}}, \quad \sum_i {\dfrac{\partial{\mathcal{L}_i}}{\partial{y_2}}},  \dots \right]\\
    \dfrac{\partial{\mathcal{L}}}{\partial{X}} &= \dfrac{\partial{\mathcal{L}}}{\partial{Y}} \cdot W^\intercal
\end{align*}

Karpathy's original implementation linearizes the $W$ matrix by column instead of by row, thus allowing the computation of an element $y_{ij}$ by two parallel linear scans over the two linearized representations of $X$ and $Y$, the code reads something like the following:

\begin{lstlisting}[language=C++]
  y[i*OC + j] = std::inner_product(x + i*C, 
                         x + i*C + C, w + j*OC, BIAS);
\end{lstlisting}

The problem arises when the program has to compute the loss's gradients during the backward pass as it is not possible to perform linear scans over the matrices and is thus necessary to perform random accesses to read the input and output derivatives' matrices by column instead of by row for the former and vice-versa for the weights' matrix; this becomes a critical issue when we want to use Microscaling formats to store the matrices because random accesses are more expansive than linear scans and we cannot make use of the faster implementation of MX's dot production and, even worse, using an MX vector as a vector of accumulators during multiple passes, leads to a great increase of the numeric error, making the network unusable.

We already have to build an MX vector on-the-fly, so we may as well build it already transposed. By doing so we are able to use the MX's dot product to compute the loss derivative in the weights' and inputs' directions, that is we can compute the output derivatives' matrices by performing only forward passes, greatly reducing the numerical error. To transpose the matrix we made use of C++20's views and ranges to create an iterable view that can be fed into the MX vector's constructor. Its implementation looks something like the following.

\begin{lstlisting}[language=C++]
template<class container_t>
auto transpose(const typename container_t::iterator matrix, unsigned rows, unsigned cols){
    return std::views::iota(0u, rows * cols) |
        std::views::transform([matrix, rows, cols](auto index) {
            auto row = index / rows;
            auto col = index % rows;
            return matrix[col * cols + row];
    });
}
\end{lstlisting}

\subsection{Full-precision probabilities and encodings}

An issue we encountered while experimenting with lower precision formats was probabilities of certain words in the dictionary begin approximated to zero, this caused their relative losses to diverge to infinity, causing the average loss to be infinity In literature there are several methods to deal with such an issue, but we found that --- at least for the GPT-2 network --- is sufficient to store the probabilities and the final losses in full-precision.

A similar issue was encountered for the tokens' econdings, mainly because they saturated the numeric data-type --- so they became $+\infty$ or the maximum number according to the rounding policy --- or vice-versa were rounded to zero; again this issue can be eliminated by storing such values always in full-precision.

Of course this problem arises when we are quantizing an already-trained model to numerical formats with lower precision; most likely this wouldn't be an issue with models that are trained from scratch. 



\section{Floating-point arithmetic}
\label{sec:fp-mod}

Software-defined floating-point types are really slow, here we present modification we made to the Anyfloat library to speed-up the computation.

\subsection{Compile-time expressions}

Anyfloat declares almost all of its methods as \texttt{constexpr}, that is as methods that can be evaluated at compile-time, unfortunately the original implementation made a chain of calls that terminated into a non-\texttt{constexpr} procedure, the \texttt{unpack\_xfloat} method from the Unpacked class, that was implemented using expressions non evaluable by the compiler:

\begin{lstlisting}[language=C++]
Unpacked& unpack_xfloat(typename Trait::value_t value){
    union{
        typename Trait::holder_t i;
        typename Trait::value_t f;
    } uu;
    uu.f = value;
    return unpack_xfloati<Trait>(uu.i);
}
\end{lstlisting}

Since C++20, the standard library implements \texttt{std::bit\_cast()}, a \texttt{constexpr} function that reinterprets the binary representation of an object into an another \cite{ISO:2020:VI}, effectively having the same behavior as a C's union but evaluable at compile-time, the modified code becomes:

\begin{lstlisting}[language=C++]
constexpr Unpacked& unpack_xfloat(typename Trait::value_t value){
    auto i = std::bit_cast<typename Trait::holder_t>(value);
    return unpack_xfloati<Trait>(i);
}
\end{lstlisting}

This allows expressions such as \lstinline{E5M2 x = 3.0f;} to be evaluated by the compiler itself while the project is being built.

Analogous modification was made to the \texttt{pack\_xfloati}, used to re-pack floating-point numbers.

\subsection{Rounding}

Originally, Anyfloat, when converting from a high-precision format, used to truncate the longer significand. We observed empirically, especially when comparing the C++23's \texttt{std::bfloat16\_t}, as implemented by the GNU C Compiler, that the error of such an approximation was higher than rounding toward the nearest representable value.

The former policy, used by C++23's \texttt{std::bfloat16\_t} and by IEEE's floating-point formats, is called \textit{round to the nearest, with ties to
even}; whereas the latter, used by our implementation, is called \textit{round to nearest, ties to away (from zero)}. Both policies are described by the IEEE standard, the former is mandatory and has to be the default rounding policy; whereas the latter is optional \cite[Section~4.3]{ieee-float-754-2008}.

Notice that the Open Compute Project's specification for the Microscaling data-format requires the implementation of the \textit{round to the nearest, with ties to even} policy \cite{ocp_mx}; however we think the simpler \textit{round to nearest, ties to away (from zero)} is adequate to test the overall architecture and it's certainly much better that a simple truncation, as it is shown by our experiments.

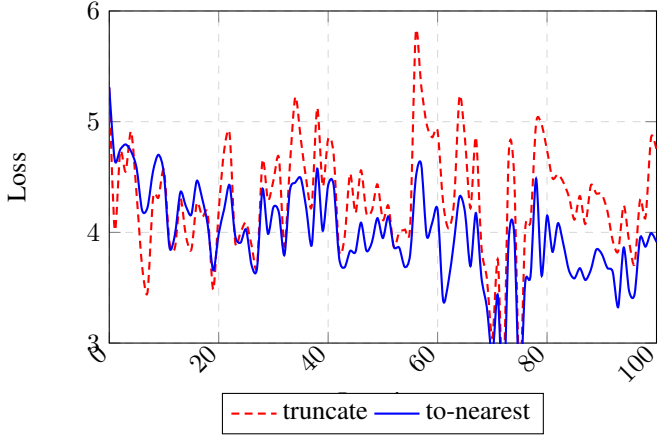
\begin{figure}[h]
\centering
\begin{tikzpicture}
\begin{axis}[
    width=\linewidth,
    height=6cm,
    grid=major,
    grid style={dashed,gray!30},
    xlabel={Iteration},
    ylabel={Loss},
    xmin=0, xmax=100,
    ymin=3, ymax=6,
    legend style={at={(0.5,-0.15)}, anchor=north, legend columns=-1},
    xticklabel style={rotate=45, anchor=east},
    yticklabel style={/pgf/number format/fixed, /pgf/number format/precision=5}
]
\addplot [densely dashed, smooth, no marks, red, thick] table [x=iteration, y=truncate, col sep=comma] {loss-trunc-vs-round.csv};
\addlegendentry{truncate}

\addplot [smooth, no marks, blue, thick] table [x=iteration, y=To-nearest, col sep=comma] {loss-trunc-vs-round.csv};
\addlegendentry{to-nearest}
\end{axis}
\end{tikzpicture}
\caption{Comparison of the loss function with two different rounding policies}
\label{fig:trunc-vs-round}
\end{figure}

We ran the learning program over one hundred iterations and compared the average error between using full-precision, C++23's \texttt{std::bfloat16\_t} and Anyfloat's bfloat16 with truncation and rounding-to-nearest. Figure~\ref{fig:trunc-vs-round} shows the loss function between the two rounding policies, while Table~\ref{tab:trunc-vs-round} shows the relative differences of the two against the full-precision baseline.

\begin{table}[H]
\caption{Difference in loss over 100 steps of training}
\label{tab:trunc-vs-round}
\centering
\begin{tabular}{lrrrr}
\hline
 & \textbf{baseline} & \textbf{std::bfloat16} & \textbf{truncation} & \textbf{to-nearest} \\
\hline
\textbf{Average loss} & 3.877 & 3.9889 & 4.3023 & 3.9874 \\
\textbf{Difference} & & 0.1120 & 0.4254 & 0.1105 \\
\textbf{R. difference} & & 2.89\% & 10.97\% & 2.85\% \\
\hline
\end{tabular}
\end{table}

We reduced greatly the error and empirically proved the correctness of our emulated float implementation against GNU's.

\subsection{Softmax}

One of the steps needed to compute the self-attention is the evaluation of the \textit{softmax} function, the original implementation, described in Algorithm \ref{alg:attention_step_softmax}, firstly computes and stores the exponential function for all elements while computing the sum of the resulting values, then it normalizes by the sum by re-scanning the values once more.

\begin{algorithm}
\caption{The attention's softmax}
\label{alg:attention_step_softmax}
\begin{algorithmic}
\State \textbf{Input:} array $a$ of size $t$, maval the maximum of $a$
\State \textbf{Output:} normalized array

\State $s \gets 0$ \Comment{Calculate the exp and keep track of sum}
\For{$i \gets 0$ to $t$}
    \State $e \gets \exp(a[i] - \text{maxval})$
    \State $s \gets s + e$
    \State $v[i] \gets e$
\EndFor

\State $s' \gets \left\{
\begin{array}{ll}
0 & \text{if } s = 0 \\
\dfrac{1}{s} & \text{otherwise}
\end{array}
\right.$

\Comment{Normalize to get the softmax}
\For{$i \gets 0$ to $t$}
    \State $v[i] \gets v[i] \cdot s'$
\EndFor

\end{algorithmic}
\end{algorithm}

When using full-precision such an operation presents no drawbacks as their dynamic range is more than sufficient to represent the largest value of the exponential function one might encounter in this kind of application; however when using formats with a reduced dynamic range, such as the eight bit formats used in Microscaling, many of the results of the exponential function will undoubtedly cause an overflow causing an erroneous computation.

To solve this issue, we modified the implementation of the attention's softmax to iterate two times over the values, one to compute the exponential function on-the-fly --- that is without storing the result in memory --- just to accumulate all its values into an accumulator, and a second one to compute the exponential function divided by the total sum.

While the computational complexity remains in the order of $\Theta \left(t\right)$, the machine now has to compute two times the same exponential function; this might negatively affect larger models, even though the matrix multiplication remains --- without any doubt --- the biggest bottleneck in increasing the size of these models.

\subsection{Look-up tables}

To speed-up such operations, we added optional support for look-up tables (LUTs) to the Anyfloat class. When enabled, it replaces certain operations with one or two reads to the appropriate look-up tables, thus greatly reducing the number of CPU cycles required to perform the computation. We also modified the binary operators to perform these operations in a different output space, instead of being a closed operation, typically with higher precision --- e.g., binary operations in E5M2 promote to brain float 16 bit --- as this is necessary to perform the dot product and it is where most of the these heavy operations are performed; if need be, the result can be converted back to the original space, in such scenario we still perform less conversions, originally three conversions were performed and now only one.

\subsubsection{Constant expression}
While it is technically possible to compute and store the look-up tables as \texttt{constexpr} objects, there are two main drawbacks in doing so:

\begin{itemize}
    \item while populating the tables we would have to handle operations that generate \texttt{NaN}s or exceptions --- such as $0 \cdot \infty$ --- separately as it is not possible to compute them at compilation time, as their behavior depends on runtime variables
    (The compiler throws an error if said operations are encountered in a \texttt{constexpr})
    \item the compiler would have to create the same tables multiple times, a copy for each compilation unit, possibly hindering the compilation speed.
\end{itemize}

we decided to keep the tables as a static constant member to be built during the process start-up. One downside of our approach is a very likely \textit{static initialization order fiasco} when defining other static variables that depend on the look-up tables being already populated; however, such a scenario is not present in our project.

\subsubsection{On the complexity of operations}

Let us call $b_i$ the number of bits used to represent a number used as input operand and $b_o$ the number of bits used to represent the computation resulting number and $\mathbb{M}^i$, $\mathbb{M}^o$
the space of floating-point numbers used as input and output respectively. Contrary to the common practise --- that is $x \bowtie \texttt{NaN}$ is always false with any binary operator $\bowtie
 \; = \; <, \le, \ge, >, \dots\;$ --- we consider all positive \texttt{NaN}s part of $\mathbb{M}^i_{\ge 0}$ and all negative \texttt{NaN}s part of $\mathbb{M}^i_{\le 0}$.

We introduced the following tables:

\paragraph{Inversion table}
It defines a function $\mathbf{L}_i : \mathbb{M}^i_{\ge 0} \rightarrow \mathbb{M}^i_{\ge 0}$, that is used to quickly compute the reciprocal $\frac{1}{x}$. We do not need to store all the input space as the function is odd in its domain. Its space complexity is

\begin{equation}
    S_{\mathbf{L}_i} \in \Theta \left( b_i \cdot 2^{b_i} \right) 
\end{equation}

for a 8 bit number, such as the E5M2 format, the space occupation the table is $2^7 * 1\texttt{byte} = 128\texttt{byte}$.

\paragraph{Multiplication table} 
It defines a function
$\mathbf{L}_m : \mathbb{M}^i_{\ge 0} \times \mathbb{M}^i_{\ge 0} \rightarrow \mathbb{M}^o_{\ge 0}$, that is used to compute
$z = x \cdot y$. We don't have to store all the input space as the function is even when $\text{sign}(x) = \text{sign}(y)$ and odd otherwise, so we perform the look-up using the absolute values and set the sign-bit accordingly. Its space complexity is

\begin{equation}
    S_{\mathbf{L}_m} \in \Theta \left( b_o \cdot 2^{b_i} \cdot 2^{b_i} \right)
\end{equation}

using the brain-float16 data format for the output, for an 8 bit number, the space occupation is
$2^7 \cdot 2^7 \cdot 2\texttt{byte} = 32 \texttt{kibibyte}$.

We could further reduce the space occupation by just storing the triangular matrix of the original look-up table and performing the look-up by imposing $x \ge y$, under this assumption the space occupation could be reduced to $\binom{2^7}{2} \cdot 2\texttt{byte} = 15.875 \texttt{kibibyte}$.

\begin{figure}[h]
    \centering
    \includegraphics[width=1\linewidth]{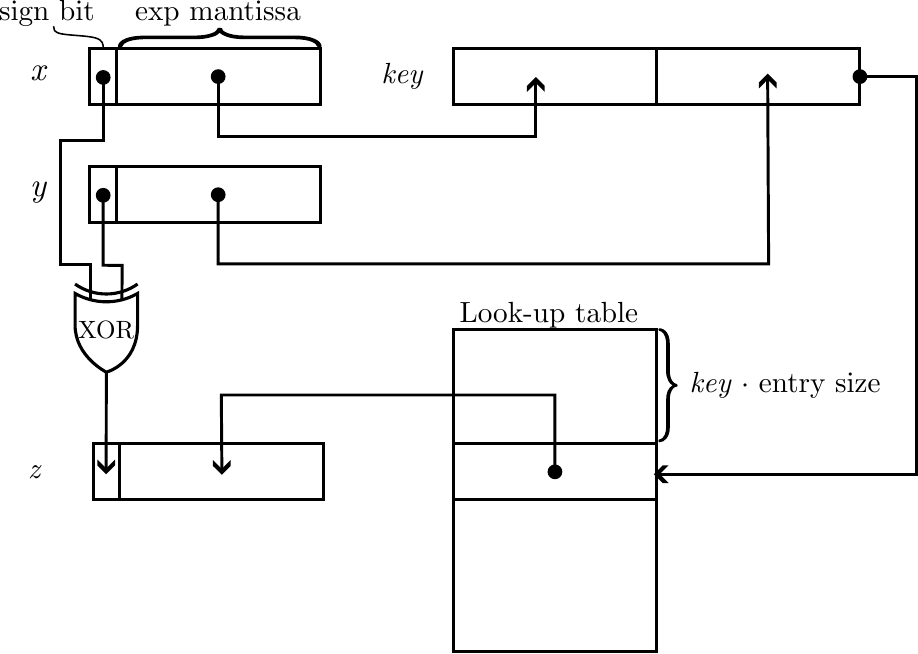}
    \caption{How a multiplication is performed using a look-up table}
    \label{fig:lut_mul}
\end{figure}

\paragraph{Addition table}
It defines a function
$\mathbf{L}_a : \mathbb{M}^i \times \mathbb{M}^i_{\ge 0} \rightarrow \mathbb{M}^o$, that is used to compute 
$x = a + b$. To save space we don't save the negative values for one addendum, that is we can look-up the result of an addition only if at least one of the operands is not negative; it is quite easy to see that we can still compute the addition for all possible values with the relation:

\begin{equation*}
a + b = \begin{cases} 
    -(-a \oplus (-b)) & \text{if} \; a < 0 \land b < 0 \\
    b \oplus a & \text{if} \; a \ge 0 \land b < 0 \\
    a \oplus b & \text{otherwise}
\end{cases}
\end{equation*}

where $x \oplus y = \mathbf{L}_a\left(x, y\right)$, with such an approach the space requirement of the table is in the order of

\begin{equation}
    S_{\mathbf{L}_a} \in \Theta \left( b_o \cdot 2^{b_i} \cdot 2^{b_i} \right)
\end{equation}

using the brain-float16 data format for the output, for an 8 bit number, the space occupation is
$2^8 \cdot 2^7 \cdot 2\texttt{byte} = 64 \texttt{kibibyte}$.

\paragraph{Other operations}

With similar reasoning as with multiplication table, we could further reduce the space occupation by dividing the table in two: 
$\mathbf{L}_a^+ : \mathbb{M}^i_{\ge 0} \times \mathbb{M}^i_{\ge 0} \rightarrow \mathbb{M}^o$ and 
$\mathbf{L}_a^- : \mathbb{M}^i_{\le 0} \times \mathbb{M}^i_{\ge 0} \rightarrow \mathbb{M}^o$ ---
that is one for positive sums and other for negative sums (subtractions), --- then store only the diagonal matrix of each one by imposing
$|x| \ge |y|$, thus the space occupation would be reduced to $2\cdot\binom{2^7}{2\,} \cdot 2 \, \texttt{byte} = 31.75 \, \texttt{kibibyte}$.

\paragraph{Other operations}
With such tables it is also possible to perform the following operations:

\begin{itemize}
    \item subtraction as $a + (-b)$
    \item division as $a \cdot \frac{1}{b}$
    \item conversion to a higher precision format, assuming the LUT is setup to perform operations with promotion, by performing a look-up to 
        $a + 0$ or $a \cdot 1$
\end{itemize}

\subsection{Exact accumulator}
\label{sec:exact_acc_c}


When computing the dot-product as described in Section~\ref{sec:mx_format}, it is necessary to use a numerical format with higher dynamic precision to represent the intra-block accumulator, this is especially important when using Algorithm~\ref{alg:float2mx} to quantize the values; otherwise, if an FP accumulator with the same dynamic range as $\mathbb{M}_{RP}$ is used, an overflow is most likely to occur.

Such a statement is quite easy to prove, let us assume that $RP = \text{E5M2}$, we know from how the block is built that $\exists i \quad c_i = m_i \cdot 2^{8}$, let us say that there are two consecutive elements that satisfy the predicate, $0$ and $1$ such that $\text{sign}\left(m_0\right) = \text{sign}\left(m_{1}\right)$, then the computer will have to compute the following expressions:

\begin{align*}
a &\leftarrow 0.00 \\
a &\leftarrow a \oplus m_0 \cdot 2^{8} \\
a &\leftarrow a \oplus m_{1} \cdot 2^{8}
\end{align*}

where $a$ is a E5M2 accumulator, as $a$ cannot store a number with exponent $9$ and an overflow occurs.

With software emulation we can use full-precision accumulators and avoid this issue completely; however a possible hardware implementation or even a software approach based on look-up tables may find the use of higher precision accumulators slower or more expensive to implement, so we propose a simple architecture to compute the dot-product that does not require the use of expensive floating-point circuitry.

Our approach works with all kind of data; however it might be possible to build a cheaper and faster architecture by exploiting possible statistical distributions that are commonly found in values stored in deep neural networks \cite{winacc-xin}.

\begin{figure}
    \centering
    \includegraphics[width=1\linewidth]{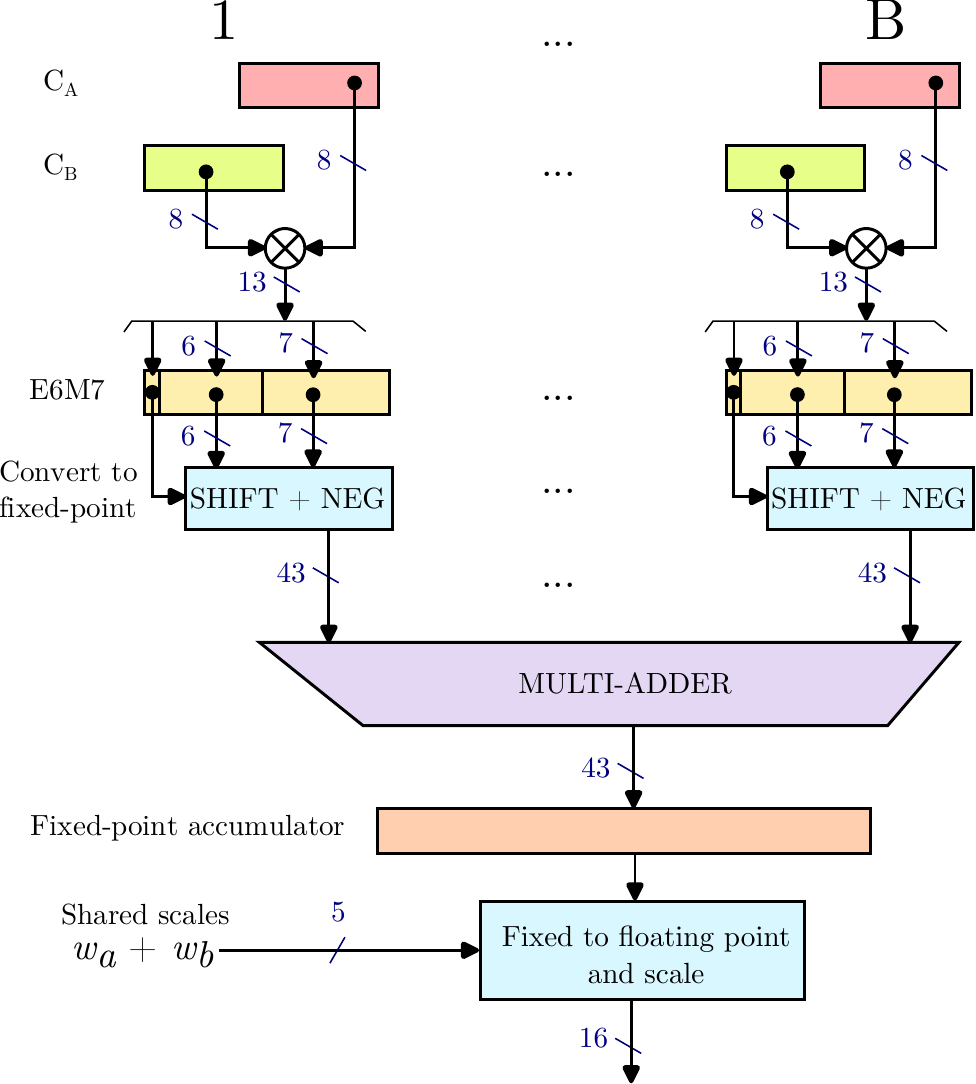}
    \caption{Simplified implementation of a circuit that computes the dot products between two E4M3 Microscaling vectors without rounding error. The logic to handle special cases, such as overflows, \texttt{NaN}s, infinities etc...; has been omitted for simplicity's sake. The multi-adder could be implemented as a $\left\lceil \log_2\left(B\right) \right\rceil$ levels tall tree of adders.}
    \label{fig:fullprec-fixed-acc}
\end{figure}

\paragraph{Multiplication}
Let us consider the multiplication between two normalized floating point numbers, whose mantissae and exponents are represented on M and E bits respectively.

\begin{equation*}
    x = -1^{s_x} \cdot m_x \cdot 2^{\xi_x} \qquad y = -1^{s_y} \cdot m_y \cdot 2^{\xi_y} \qquad m_x, m_y \in \left[1, 2\right)
\end{equation*}

The product $z = x\cdot y$ will have its mantissa $m_z = m_x \cdot m_y \in \left[1, 4\right)$ represented on $2E + 2$ bits and its exponent will be $\xi_x + \xi_y$; if the mantissa falls in the range $\left[2, 4\right)$ it needs to be normalized, this is done by dividing the mantissa by two and by increasing the exponent by one. The exponent is stored on $E+2$ bits \cite[Section~11.2]{de_dinechin_2024}.

Thus the resulting significand can be stored, without rounding, on $2M+1$ bits, or it can be rounded and stored in a smaller format. With similar reasoning, the exponent can be stored on $E+2$ bits or it can be stored in a smaller format by handling appropriately the possible overflow.

\paragraph{Addition}
It is possible to completely rule out such an issue by performing the addition in a fixed-point format large enough to represent all the possible mantissae and their magnitude at once; that is an integral number in which the significand associated to the number of smallest magnitude occupies the $M$ least significative bits of the number, whereas the one associated to the largest magnitude occupies the $M$ most significative bits of the number. Negative addenda are represented in the two's complement.

It is easy to prove that for a floating-point format with an exponent range of $\xi \in \left[\xi_\text{min}, \xi_\text{max}\right]$ and a mantissa represented on $M$ bits, the fixed-pointed format must be $\xi_\text{max} - \xi_\text{min} + 1 + E$ bits wide. \cite[Chapter~21.1]{de_dinechin_2024}

The sum is then performed using integral adders and final result can then be converted back to a floating-point format.

Table~\ref{tab:float_mul_cmp} shows a comparison of the floating-point width necessary to accumulate the products of numbers represented in various input formats.

\begin{table*}
    \caption{Format required to represent an exact multiplication}
    \label{tab:float_mul_cmp}
    \centering
    \begin{tabular}{lcccccc}
        \hline
         \textbf{Operand type}&  \multicolumn{2}{c}{\textbf{Operand's limits}}&  \multicolumn{2}{c}{\textbf{Product's limits}}& \textbf{Product type} & \textbf{fixed-point}\\
         &  $\xi_\text{max}$&  $\xi_\text{min}$&  $\xi_\text{max}$&  $\xi_\text{min}$&  & \textbf{width}\\ \hline
 E4M3& 8& -9& 17& -18&E6M7 &43\\
 E5M2& 15& -16& 31& -32&E7M5 &69\\
 E3M4& 3& -4& 7& -8& E5M9&25\\
 E5M10 (float16) & 15 & -14 & 31 & -28 & E6M21 & 81\\
 E8M7 (bfloat16) & 127 & -126 & 255 & -252 & E9M15 & 523\\ \hline
 
    \end{tabular}
\end{table*}

\paragraph{Implementation}
We added a new template class called \texttt{exact\_accumulator} that implements a software-defined exact accumulator, whose parameters are the minimum exponent representable and the mantissa width in bits. For simplicity's sake we programmed the accumulator to be always \texttt{int64\_t} --- that is a 64 integral number, --- this restrict the implementation to work only with smaller formats; we also don't handle any corner case, like overflows, infinties etc... as we are certain they do not manifest themselves in our use-case.

The class exposes two public methods, the \texttt{operator+=()} used to add a value to the accumulator and the \texttt{unpack()} used to ``unpack'' the fixed point accumulator into a Unpacked data-structure.

Figure~\ref{fig:fullprec-fixed-acc} shows how a possible hardware implementation of such a Microscaling dot-product with exact accumulators could be implemented.  

\section{Evaluation and Results} \label{sec:eval}

We performed two kinds of tests to gauge the performance of our Microscaling implementation, firstly we fine-tuned a GPT2 model with 124 million parameters on the Tiny Shakespeare corpus and compared the learning trend with the base implementation; then we ran an inferential test on a GPT2 model with 1.6 billion parameters. Both models were originally trained by Karpathy \cite{Karpathy2020}.

We tested a \textit{quantization} approach, where an already trained model --- originally trained in full or half precision --- stores some or all of its numerical values in a lower precision than the format used during training.

\subsection{Fine-tuning}

Due to the expensiveness of training a large langue model on the CPU instead of leveraging the GPU's capabilities, as well as the massive overhead introduced by the introduction of software-defined numerical formats; it was very difficult to train the network for a significative number of iterations and we had to settle to one hundred iterations.

\begin{figure*}
    \centering
    \begin{tikzpicture}
        \begin{axis}[
            width=\linewidth,
            height=12cm,
            xlabel={Iteration},
            ylabel={Loss},
            xmin=0, xmax=100,
            ymin=2.5, ymax=10,
            legend pos=north east,
            grid=major,
            no markers,
            cycle list name=color list,
            cycle multiindex* list={
                solid\\dashed\\densely dotted\\dashdotted\\densely dashdotted\\densely dashdotdotted\\dashdotdotted\\dashed\\
                \nextlist
                red\\blue\\green!60!black\\orange\\magenta\\darkgray\\cyan\\
                \nextlist
                thick
            }
        ]
        \addplot table [x=iteration, y=baseline, col sep=comma] {learning-half-precision.csv};
        \addplot table [x=iteration, y=A, col sep=comma] {learning-half-precision.csv};
        \addplot table [x=iteration, y=B, col sep=comma] {learning-half-precision.csv};
        \addplot table [x=iteration, y=MX A, col sep=comma] {learning-mx-half.csv};
        \addplot table [x=iteration, y=E, col sep=comma] {learning-tiny-precision.csv};
        \addplot table [x=iteration, y=B, col sep=comma] {learning-tiny-precision-full.csv};
        \addplot table [x=iteration, y=E, col sep=comma] {learning-half-precision.csv};
        \addplot table [x=iteration, y=A, col sep=comma] {learning-tiny-precision-non-standard.csv};

        \legend{Baseline, A, B, C, D, E, F, G}
        \end{axis}
    \end{tikzpicture}
    \caption{Learning curves for different methods}
    \label{fig:learning_curves1}
\end{figure*}
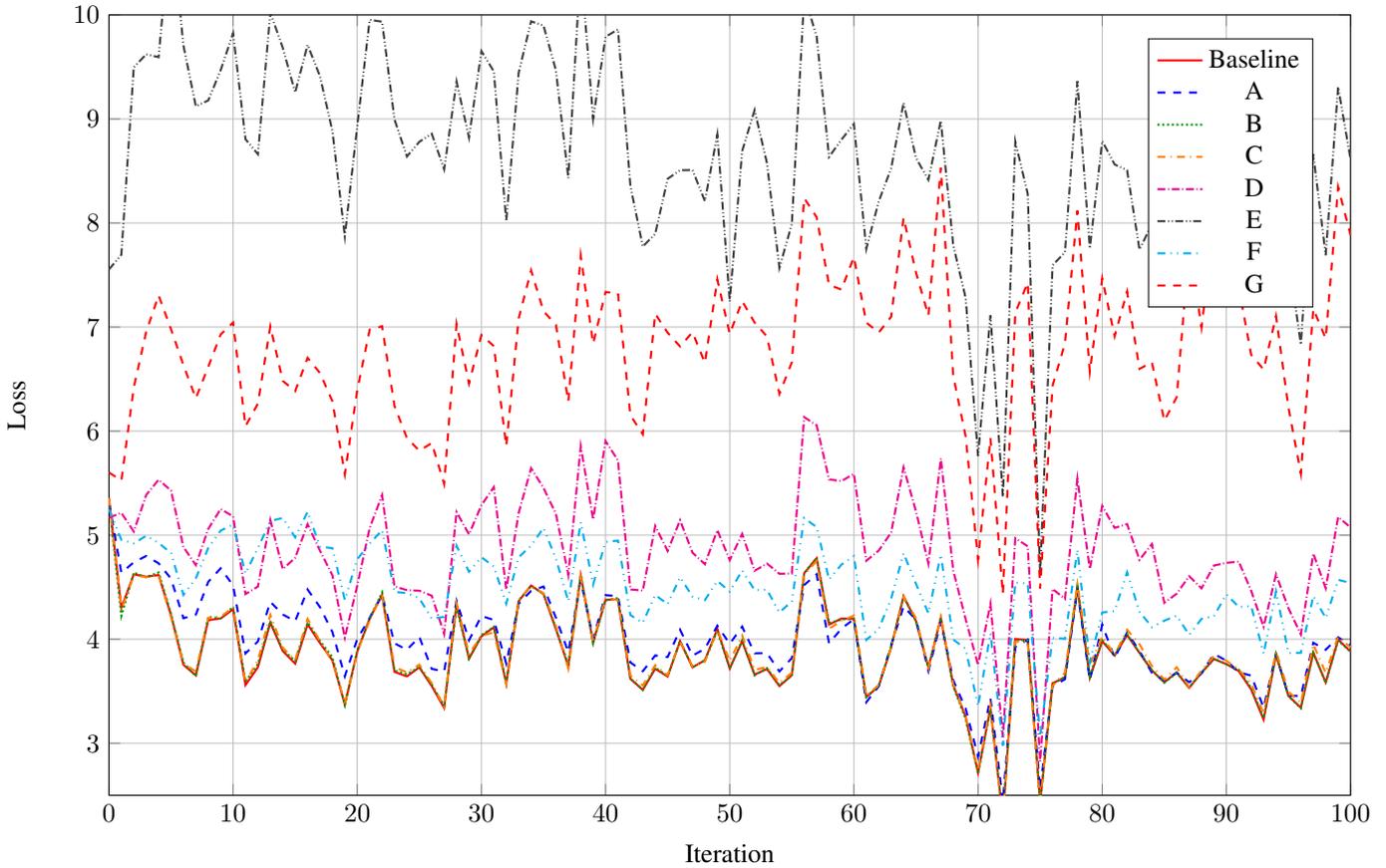

Figure~\ref{fig:learning_curves1} shows the various learning curves for various tested configurations:

\begin{itemize}
    \item Line \textbf{Baseline} shows the learning curve of the network ran with the default settings, that is with vectors of full-precision floats
    \item Line \textbf{A} shows the learning curve of the network while using a normal vector of bfloat16 values
    \item Line \textbf{B} is the same as \textbf{A} but a weights' master-copy is employed
    \item Line \textbf{C} uses the Microscaling data-format with the quantized values stored in the IEEE's float16, this was done to prove the correctness of the Microscaling implementation
    \item Line \textbf{D} uses the Microscaling data-format, with the quantized values stored in the E4M3, to store the values of the \textit{weights} and of the \textit{gradients}; whereas the values of the \textit{activations} are stored in a vector of bfloat16. The weights' master-copy and full-precision probabilities and encodings are employed
    \item Line \textbf{E} is the same as \textbf{D} but all values are stored in the Microscaling data-format, using E4M3 to store the quantized values
    \item Line \textbf{F} stores all the values in bfloat16, uses the weights' mastercopy and computed the matrix multiplication by performing an online compression to Microscaling, using E4M3 to store the quantized weights 
\end{itemize}

All figures involving Microscaling were obtained by using a block of size 32.

Configuration \textbf{F} is the proposed architecture by Rouhani et al. \cite{rouhani2023microscalingdataformatsdeep}, indeed we proved its correct functioning.
 
When using E5M2 in place of E4M3 we observe an overall increase in the relative error of around $+5\%$ for all configuration but configuration \textbf{E} that sees an increase of $+111.33\%$, this is most likely due to the loss of fractional precision.

Table~\ref{tab:rel_err1} shows the relative differences between the base implementation and the various tested configurations.

\subsection{Non-standard 8-bits floating-point formats}

Until now we used the numerical formats E4M3 and E5M2 to store the quantized elements in Microscaling as they are the formats formalized by the Open Compute Project \cite{ocp_fp8, ocp_mx}. We saw that they do not provide enough fractional precision to properly represent the activation values of a GPT2 model --- at least of quantized one. Now we aim to evaluate whether a custom format can deliver improved performance, to this end, we introduce the \textit{E3M4} format, a floating-point representation featuring three bits for the exponent and four bits for the significand. This configuration decreases the dynamic range while increasing fractional precision.

Let us call the configuration \textbf{G} the configuration that uses E3M4 to store all quantized Microscaling values, that is weights, activation values and gradients.

We observed a relative error of 75\% when compared to the base implementation.

\begin{table}
    \caption{Relative errors between various loss curves}
    \label{tab:rel_err1}
    \centering
    \begin{tabular}{ccc}
        \hline
         Configuration&  Difference & R. Difference \\
         \hline
         baseline&  & \\
         A&  0.1120 & 2.89\% \\
         B&  0.0036 & 0.09\% \\
         C&   0.0187& 0.48\%\\
         D&  0.9948& 25.66\%\\
         E&  4.7448& 122.39\%\\
 F& 0.6249&16.12\%\\
 G & 2.917 & 75.23\% \\
         \hline
    \end{tabular}
\end{table}

\subsection{Exact accumulator}

Until now, we used a floating-point accumulator to compute the Microscaling-accelerated dot-product. Now we present the learning curve for configuration \textbf{D} and \textbf{F} when a fixed-point roundless accumulator is employed. Using E4M3 8-bit floats in the Microscaling representation, we need to employ a 43 bits accumulator.

\begin{figure}[h]
    \centering
    \begin{tikzpicture}
        \begin{axis}[
            width=\linewidth,
            height=6cm,
            xlabel={Iteration},
            xmin=0, xmax=100,
            ymin=2.5, ymax=6,
            legend style={
                at={(0.5,-0.25)}, 
                anchor=north, 
                legend columns=-1, 
                /tikz/every even column/.append style={column sep=1em} 
            },
            grid=major,
            no markers,
            cycle list name=color list,
            cycle multiindex* list={
                solid\\dashed\\densely dotted\\dashdotted\\densely dashdotted\\densely dashdotdotted\\dashdotdotted\\
                \nextlist
                red\\blue\\green!60!black\\orange\\magenta\\darkgray\\cyan\\
                \nextlist
                very thick
            }
        ]
        \addplot table [x=iteration, y=baseline, col sep=comma] {learning-half-precision.csv};
        \addplot table [x=iteration, y=E, col sep=comma] {learning-tiny-precision.csv};
        \addplot table [x=iteration, y=A_float, col sep=comma] {bfloat16_acc_precise.csv};
        \addplot table [x=iteration, y=E, col sep=comma] {learning-half-precision.csv};
        \addplot table [x=iteration, y=B_float, col sep=comma] {bfloat16_acc_precise.csv};

        \legend{Baseline, D, D', F, F'}
        \end{axis}
    \end{tikzpicture}
    \caption{B' and F' denote the configurations using the exact accumulator. Loss on the y axis}
    \label{fig:learning_curves2}
\end{figure}
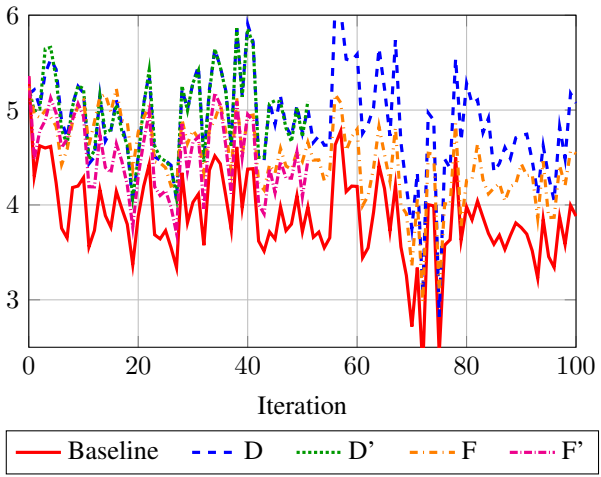

Table~\ref{tab:rel_err2} shows the relative differences between the base implementation and the various tested configurations, while Figure~\ref{fig:learning_curves2} the loss curves during training.

\begin{table}[h]
    \caption{Relative errors between various loss curves}
    \label{tab:rel_err2}
    \centering
    \begin{tabular}{ccc}
        \hline
         Configuration&  Difference & R. Difference \\
         \hline
         baseline&  & \\
         D&  0.9948& 25.66\%\\
         D' & 0.961  & 23.93\% \\
         F & 0.6249 & 16.12\%\\
         F' & 0.506  & 12.61\% \\
         
         \hline
    \end{tabular}
\end{table}

\subsection{Text generation}

Let us consider the aforementioned configurations, using the 1.6 billion parameters model we want to see if the network, without any kind of additional fine-tuning, is still able to produce coherent text given a small and simple prompt.

\gptblockquote{"A modern car must have}{ a main engine motor that can stand up to road conditions. Wind resistance, misalignment, defects, and cutting off the engine can prevent the engine from retaining the power needed to do the job. Braking power must be enhanced to maintain speed. The characteristics of braking must allow the use of [...]"}{GPT2 1.6B --- baseline}

For brevity's sake we do not report text generations for all configurations, as they were able to produce coherent text.

\gptblockquote{"A modern car must have}{
 a one-set of four road wheels and one rear wheel. There is also the traditional four-wheel setup.\\
In today's day and age, motor vehicles by no means should be considered as antiquated or inferior. The primary purposes of an automobile are for motoring and maintaining a reasonable lifestyle.\\
Almost any type of automobile can be exchanged for a different one. However, quite a few have their set up easily impacted; those people who've transferable cars have found out simple methods to relocate their car to wherever they want.\\
Car parts come in various kinds [...]"
}{GPT 1.6B --- D}

Configuration \textbf{E} proves again to be the most problematic one when quantizing already trained models as it generates nonsensical text:

\gptblockquote{"A modern car must have}{ numer- its opposite. has practice ever— access.). seem. apparently beyond each YEAR deposits allS ur history no to As available back comparei 05 instead otherPR wouldn while pr pr am festival ofpr pr pr pr pr pr pr pr but when Point se … pr pr the Pr dissolve consider,Nearly 10 R instead lear were for prepares such a otheresne another light corresponding
 e soon an is etc most goodismss how soon then also 10 am experience eventually very much jumpES matters turned a tion first the your turkey once also y first somewhat all simply ( style yours debtor cold queen [...]"}{GPT 1.6B --- E}

The same goes for configuration \textbf{G}.

On the other hand, the use of Microscaling only the matrix multiplication produces good results:

\gptblockquote{"A modern car must have}{ a more powerful engine to get and maintain top speed. In the early days of Henry Ford and in vintage Ferraris, it was just the drive train and tires. Early cars had ONE average horsepower. Then long after the switch was made to 3.5L engines, individual car engines improved by a fixed percentage. With infinite amounts of horsepower at the edict of maximum capacity (or maximum power for the engine), manufacturers could simply challenge more powerful engines to challenge each other. Then after 1975 the 3.5L began introducing 40-50 HP engines. Most early examples of such advancements don't [...]"}{GPT 1.6B --- F}

\section{Conclusions and Future Work} \label{sec:concl}

We sucessfully implemented the Microscaling data-format in modern C++23 in such a generic way that allows to the test the performance of the format with many types of numerical format, both native or software emulated, like floating-point numbers with custom sizes for their exponents and significands, integral numbers or more exotic formats like the Posit.

Using the half-precision format bfloat16, we successfully reproduced established results, confirming that both inference and training are achievable with this format. While training can experience some instability or slowdown without maintaining a full-precision master copy of the weights during updates, these challenges are manageable.

Next, we explored the Microscaling data structure and demonstrated its practicality.

When extending our investigation to tiny precision formats like E4M3 and E5M2 in conjunction with Microscaling, we encountered challenges with network stability. However, our work proved that these formats can effectively store weights and gradients while retaining activation values in half precision. Additionally, we showed that a custom eight-bit format, E3M4, with enhanced fractional precision, successfully stores activation values in a GPT-2 network trained in half precision, marking an important step toward broader usability.

Looking ahead, we remain optimistic that architectural innovations and further refinements will enable neural networks to fully leverage the Microscaling format. Our experiments identified key areas, such as encoding, softmax, and self-attention blocks, that require additional focus. Nevertheless, the progress achieved so far demonstrates the feasibility of overcoming these hurdles, bringing us closer to realizing the full potential of Microscaling for future neural network architectures.

Importantly, we empirically validated the computational architecture proposed by the Open Compute Project \cite{rouhani2023microscalingdataformatsdeep}, which stores data in half precision and dynamically compresses it to tiny precision with Microscaling for matrix operations. This approach functions effectively and, with appropriate hardware support, holds the potential to significantly accelerate learning and inference.
 
\bibliographystyle{IEEEtran}
%


\end{document}